\documentclass[lettersize,journal]{IEEEtran}
\usepackage{amsmath,amsfonts}
\usepackage{algorithmic}
\usepackage[ruled,linesnumbered]{algorithm2e}
\usepackage{soul, color, xcolor}
\usepackage{array}
\usepackage[caption=false,font=normalsize,labelfont=sf,textfont=sf]{subfig}
\usepackage{float}
\usepackage{booktabs}
\usepackage{textcomp}
\usepackage{stfloats}
\usepackage{url}
\usepackage{verbatim}
\usepackage{graphicx}
\def\BibTeX{{\rm B\kern-.05em{\sc i\kern-.025em b}\kern-.08em
    T\kern-.1667em\lower.7ex\hbox{E}\kern-.125emX}}
\usepackage{balance}
\begin{document}
\renewcommand{\arraystretch}{1.33} 

\title{Near-field Beam Training for Extremely Large-scale MIMO Based on Deep Learning}
\author{Jiali Nie,   
        Yuanhao Cui,~\IEEEmembership{Member,~IEEE,}
        Zhaohui Yang,~\IEEEmembership{Member,~IEEE,}\\
        Weijie Yuan,~\IEEEmembership{Member,~IEEE,}
        Xiaojun Jing~\IEEEmembership{Member,~IEEE}
\thanks{Manuscript received 30 April 2024; revised 11 June 2024; accepted xx. Date of publication xx. An early vision of this paper was presented at the 2024 IEEE Global Communications Conference. \textit{(Corresponding author: Yuanhao Cui.)}}
\thanks{Jiali Nie, Yuanhao Cui, and Xiaojun Jing are with the School of Information and Communication Engineering, Beijing University of Posts and Telecommunications, Beijing 100876, China (e-mail: niejl@bupt.edu.cn; cuiyuanhao@bupt.edu.cn; jxiaojun@bupt.edu.cn).}%
\thanks{Zhaohui Yang is with the College of Information Science and Electronic
Engineering, Zhejiang University, Hangzhou, Zhejiang 310027, China (e-mail:
yang\_zhaohui@zju.edu.cn).}
\thanks{Weijie Yuan is with the Department of Electrical and Electronic Engineering, Southern University of Science and Technology, Shenzhen 518055, China (e-mail: yuanwj@sustech.edu.cn).}
}

\markboth{IEEE Transactions on Mobile Computing,~Vol.~x, No.~x, xx~xx}%
{How to Use the IEEEtran \LaTeX \ Templates}

\maketitle

\begin{abstract}
Extremely Large-scale Array (ELAA) is considered a frontier technology for future communication systems, pivotal in improving wireless systems' rate and spectral efficiency. As ELAA employs a multitude of antennas operating at higher frequencies, users are typically situated in the near-field region where the spherical wavefront propagates. The near-field beam training in ELAA requires both angle and distance information, which inevitably leads to a significant increase in the beam training overhead. To address this problem, we propose a near-field beam training method based on deep learning. We use a convolutional neural network (CNN) to efficiently learn channel characteristics from historical data by strategically selecting padding and kernel sizes. The negative value of the user average achievable rate is utilized as the loss function to optimize the beamformer. This method maximizes multi-user networks' achievable rate without predefined beam codebooks. Upon deployment, the model requires solely the pre-estimated channel state information (CSI) to derive the optimal beamforming vector. The simulation results demonstrate that the proposed scheme achieves a more stable beamforming gain and significantly improves performance compared to the traditional beam training method. Furthermore, owing to the inherent traits of deep learning methodologies, this approach substantially diminishes the near-field beam training overhead.

\end{abstract}


\section{Introduction}

\IEEEPARstart{T}{he} sixth-generation (6G) wireless communication networks are anticipated to deliver heightened spectral efficiency, reduced power consumption, and increased antenna numbers and densities to support faster and more reliable data transmission for future intelligent systems, Internet of Things, and other emerging applications \cite{SaadNetwork2020, CuiNetwork2021, 6GHuawei2022}. The deployment of extremely large-scale antenna arrays (ELAA) is crucial to achieving the key performance indicators (KPIs) of 6G networks \cite{LuTWC2021, CuiJSAC2023}. Particularly in the transition high-frequency bands such as millimeter wave (mmWave) and terahertz (THz) bands, ELAA is envisioned to be widely deployed in base stations (BS) to mitigate the significant path loss inherent in highly focused beamforming \cite{ZhangVTM2019, LiuSPM2023, CuiJSTSP2024}.

However, as carrier frequencies and antenna numbers increase, the Rayleigh distance in communication systems extends significantly, reaching tens to hundreds of meters \cite{SelvanAPM2017}. For instance, in an ELAA with a 1-meter aperture operating at $60$ GHz, the Rayleigh distance can reach $400$ meters. This expansion results in dense and complex interactions among antennas, which define the near-field signal propagation characteristics. Within the Rayleigh distance, signal propagation shifts from plane waves to spherical waves \cite{LiCL2022, PoongodiSTCR2023}. 

Near-field communications present distinct advantages in enhancing coverage and signal quality, especially in high-density environments such as indoor spaces or dense urban areas. By leveraging the unique propagation characteristics of the near field, flexible beam focusing becomes possible, enabling the precise localization of beam energy onto specific spatial positions rather than specific angles as in traditional far-field communication paradigms \cite{Liu2023}. Despite these benefits, near-field communications encounter substantial challenges in channel modeling, channel estimation, and beam training due to the inapplicability of schemes designed for far-field scenarios. Specifically, near-field beam training for ELAA requires beam codebooks incorporating both angle and distance sampling, necessitating a larger number of candidate beams and increasing the complexity and cost associated with beam training \cite{ZhangWCL2022}.

Researchers have explored various approaches to optimize beam training in the near field. Authors of \cite{XieWCL2023} proposed a near-field codebook design based on a uniform circular array (UCA), optimizing the beam pattern of the UCA to cover the user's area. Despite \cite{WeiCC2022} introducing near-field extremely large-scale reconfigurable intelligent surface (XL-RIS) codebooks and hierarchical training schemes to reduce pilot overhead, the required pilot overhead remains excessive and susceptible to noise. Deep learning technology is widely used in wireless communication scheme design due to its powerful feature extraction and learning capabilities. By conducting offline training and online deployment of deep learning models for near-field beam training, computational overhead can be substantially minimized, thus aligning with the requirements of practical applications. Wang et al. \cite{LiuCL2022} proposed a deep learning-based beam training scheme that utilizes received signals corresponding to far-field wide beams to estimate optimal near-field codewords. Jiang et al. \cite{JiangCL2023} improved beam prediction performance by training a deep neural network (DNN) on near-field codebooks containing both angle and distance information to jointly predict optimal angles and distances. However, these methods require extensive codebook design and maintenance, and their performance is constrained by specific codebook designs. To our knowledge, existing beam training approaches either fail to address the excessive pilot overhead in near-field communication systems or suffer from limited adaptability and flexibility due to reliance on predefined beam codebooks. Therefore, developing a low-overhead and high adaptability near-field beam training method remains a promising research direction.

To reduce the overhead of beam training, we propose a novel near-field beam training method based on deep learning. By training a convolutional neural network (CNN) on historical data, our method aims to dynamically predict optimal beamforming parameters based on real-time channel conditions. This approach not only circumvents the limitations of traditional beam codebooks but also enhances adaptability and efficiency in optimizing beam patterns for ELAA systems. The main contributions of this paper are as follows.

\begin{itemize}
\item We propose an ELAA near-field beam training algorithm based on deep learning. Diverging from the conventional methods that concatenate real and imaginary parts into a one-dimensional vector, our approach utilizes vertical concatenation to effectively extract complex information. This method preserves the correlation information between the real and imaginary parts (phase) through strategically designed padding and convolution kernel sizes.

\item We develop a codebook-free beam training scheme. Regarding the amplitude constraints of beamforming vectors, our innovative application of Euler's formula presents an elegant solution. The output layer function is designed to output the direction of the beamforming vector, which is then converted into numerical values using Euler's formula, ensuring adherence to the specified modulus of the beamforming vector. The negative value of the average achievable rate is defined as the loss function. By minimizing this loss function, the neural network is guided to increase the average achievable rate of the system.

\item Extensive simulations confirm the effectiveness of the proposed beam training scheme. Compared to traditional beam training methods, our approach significantly improves communication rates and reduces computational complexity, highlighting its practical benefits and enhanced performance.
\end{itemize}

\textit{Organization:} The structure of this paper is outlined as follows. In section II, we review related work on near-field communications and beam training. In section III, we introduce the characteristics of near-field propagation, near-field channel models, and near-field beamforming. Subsequently, section IV provides an in-depth presentation of the near-field beamforming method based on deep learning. In section V, we provide the experimental setup and performance analysis of the proposed scheme. Finally, section VI provides a summary of the work.

\textit{Notations:} Unless otherwise indicated, matrices are represented by bold capital letters (i.e. $\mathbf{C}$), vectors are donated by bold lowercase letters (i.e. $\mathbf{h}$), and scalars are donated by plain font (i.e. $d$). $\mathbf{h}_k$ denotes the $k$-th element of the vector $\mathbf{h}$. $\mathbf{w}^H$ denotes the conjugate transpose of the vector $\mathbf{w}$. $\mathcal{C N}\left(\mu, \sigma^{2}_n \right)$ denotes a complex Gaussian random distribution with mean $\mu$ and covariance $\sigma^{2}_n$.

\subsection{Related works}

\section{Related works}
In this section, we review related work on near-field communications and beam training.

\subsection{Near-field Communications}
Near-field communications utilize ELAA to improve the performance of wireless systems. A larger antenna size and array aperture of ELAA will bring higher beamforming gains. However, it will also lead to increased precoding complexity, more challenging beam management, and growing power consumption \cite{LiuOJCOMS2023, LuIOTJ2024, CuiMCOM2023}. Currently, numerous researchers have carried out research on ELAA near-field communications. Existing massive multiple-input multiple-output (mMIMO) systems typically operate under narrowband assumptions, with beamforming and antenna spacing tailored to center frequencies \cite{PayamiACCESS2019}. However, the traditional beam training architecture in near-field wideband systems may cause beam deflection and potentially even beam splitting \cite{MyersTWC2022}.
Furthermore, concerning codebook design, the near-field channel no longer has the sparsity in the angle domain, which makes the far-field-based orthogonal angle domain Discrete Fourier transform (DFT) codebook no longer applicable \cite{SlepianBSTJ1978}. The near-field array response vector is determined not only by the direction but also by the distance of the BS and the UE. This will greatly increase the complexity of near-field codebook design \cite{ZhangMCOM2023}.

The electromagnetic propagation of the near-field region also brings new opportunities for 6G wireless communications. Near-field communications can leverage enhanced range resolution to serve multiple UEs simultaneously \cite{ZhangTWC2022}. This feature provides additional degrees of freedom (DoF) to enhance system capacity and reduce inter-user interference \cite{DecarliAccess2021}. Near-field communications also have great potential applications in perception and security. The authors of \cite{WangCL2023} proposed a near-field integrated sensing and communications (ISAC) framework, which provides additional distance dimensions for both sensing and communications. The inclusion of an extra distance dimension in near-field ISAC leads to performance improvements compared to far-field ISAC. Moreover, \cite{ZuoGC2023} proposed the concept of near-field non-orthogonal multiple access (NF-NOMA) to enable continuous interference elimination from far to near focusing beamformers at specific locations. Near-field communications' ability to discern distance differences between UEs can significantly enhance physical layer security and communication confidentiality \cite{ZhangTVT2024}.

\subsection{Traditional Far-field Beam Training}
Traditional beam training typically involves polit training and beam alignment. In the pilot training phase, the transmitter sends a series of pilot signals covering different beam directions, and the received signals are adopted to estimate channel state information (CSI) \cite{WuTVT2020}. During the beam alignment phase, the receiver feedbacks estimated information to the transmitter, prompting adjustments to the beamformers to optimize signal focus on the receiver. Many works have been done on far-field beam training and codebook design to improve beam training accuracy and reduce overhead. Specifically, to address the high power consumption associated with fully digital beamforming, \cite{SohrabiJSTSP2016} developed a heuristic hybrid beamforming scheme that achieves performance levels close to those of full digital beamforming. An efficient hierarchical codebook was developed using sub-array and deactivated antenna technology in \cite{XiaoTWC2016}. It utilizes the results of beam training from the previous layer as the code words for the current layer, resulting in a substantial reduction in beam training overhead. An enhanced layered codebook beam training scheme is proposed in \cite{YuAccess2020} to minimize propagation errors between adjacent beams. Based on the beam pattern, the power threshold of the received signal and the misalignment probability of the beam are analyzed to improve the accuracy of the beam alignment significantly.

However, the far-field beamforming method presents challenges when dealing with near-field beams because the near-field UE channel is not only affected by the angle of departure/angle of arrival (AoD / AoA) but also by the spatial distance. Using far-field beamforming techniques for near-field training can lead to an energy diffusion effect, i.e., the beam energy directed at a specific angle may disperse to multiple angles. Directly applying far-field beam training methods to near-field scenarios may cause significant performance degradation.

\subsection{Near-field Beam Training}
To address the above issue, extensive research has been conducted on near-field beamforming techniques. Specifically, \cite{CuiTCOM2022} proposed a near-field codebook that simultaneously extracts angle and distance information. This approach reduces the codebook size to a certain extent by designing a polar transformation matrix for uniform sampling in the angle domain and non-uniform sampling in the distance domain. A two-stage layered near-field beam training method was introduced in \cite{WuTVT2023}. In the initial stage, the traditional far-field codebook is employed to conduct a rough search for candidate UE's directions in the angle domain. Subsequently, in the second stage, a specially designed polar codebook is utilized to refine the search for UE's distances within the candidate angle. Beam splitting induced by wideband ELAA was ingeniously utilized in \cite{CuiTWC2023} for beam search, achieving efficient near-field beam training with minimal overhead. The proposed method employs time delay (TD) beamforming to regulate the near-field beam splitting, enabling the search for the optimal angle through frequency division and the optimal distance ring through time division. Furthermore, \cite{LiuTCOM2023} proposed two deep learning-based near-field beam training schemes, which employ deep residual networks to determine the optimal near-field RIS codeword.

Nonetheless, the escalation of antenna size and codebook complexity exacerbates the overhead of beam training in near-field communications, as previously mentioned. Therefore, it is imperative to devise a low-overhead algorithm to mitigate the challenges posed by ELAA communication systems.

\section{Near-field communications system model}
\subsection{Near-field Channel Model}
\begin{figure}[!t]
\centering
\includegraphics[width=3in]{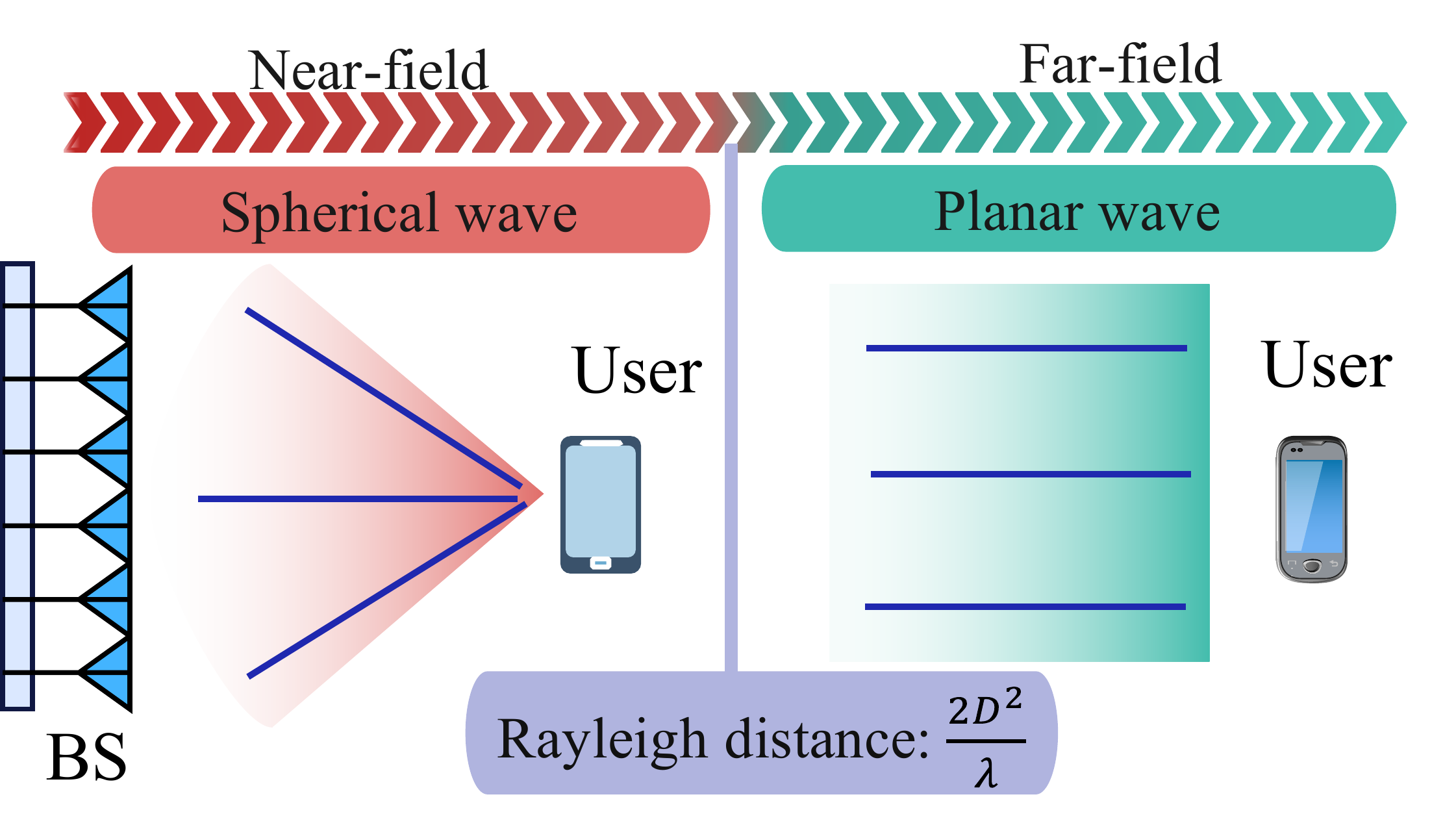}
\caption{Near-field wavefront and far-field wavefront comparison}
\label{fig1}
\end{figure}

In wireless communications, we typically delineate the division between the far field and near field using the Rayleigh distance expressed by the following \cite{Kraus2002}.
\begin{equation}
\label{eq1}
RD = \frac{2D^2}{\lambda_c}=\frac{2D^2f_c}{C},
\end{equation}
where $D$ represents the antenna diameter, $\lambda_c$ denotes the wavelength, $f_c$ is the carrier frequency, and $C$ is the rate of light. As shown in Fig.~\ref{fig1}, when the distance between the receiver and the transmitter is less than the Rayleigh distance, the signal is considered to be near-field spherical wavefront propagation.

As indicated in Table~\ref{t1}, with an antenna aperture of $0.5$ meters and a carrier center frequency of $50$ MHz, any region beyond $0.09$ meters from the BS is categorized as the far-field. With the widespread adoption of millimeter wave bands for 5G, especially when combined with large antenna arrays, near-field distances can reach tens to hundreds of meters. When the carrier frequency is $60$ GHz, and the antenna diameter is $D=0.5$ meters, any user equipment (UE) within a distance less than $100$ meters from the BS is considered in the near-field region.

\begin{table}[H]
\caption{Variation in the near-field region with carrier center frequencies.}
\centering
\begin{tabular}{c|c|c}
\toprule
Antenna diameter $(D)$&frequency $(f_c)$&Near-field Boundary \\
\midrule
0.5 m& 50 MHz          &0.09 m\\
0.5 m& 1000 MHz        &1.7 m\\
0.5 m& 5 GHz(802.11a)  &8.3 m\\
0.5 m& 28 GHz          &47 m\\
0.5 m& 60 GHz(802.11ay)& 100 m\\
\bottomrule
\end{tabular}
\label{t1}
\end{table}

In this paper, we investigate downlink beam training for extremely large-scale multiple-input-multiple-output (XL-MIMO) systems. The BS is equipped with a uniform linear array (ULA) to serve single antenna UEs. We assume that the BS has $N$ antennas with an antenna spacing of $d=\lambda_c/2$. Given the predominant reliance of mmWave communications on line-of-sight (LoS) link, we focus on beam training for near-field LoS channel of multi-user systems.

\begin{figure}[!t]
\centering
\includegraphics[width=3.5in]{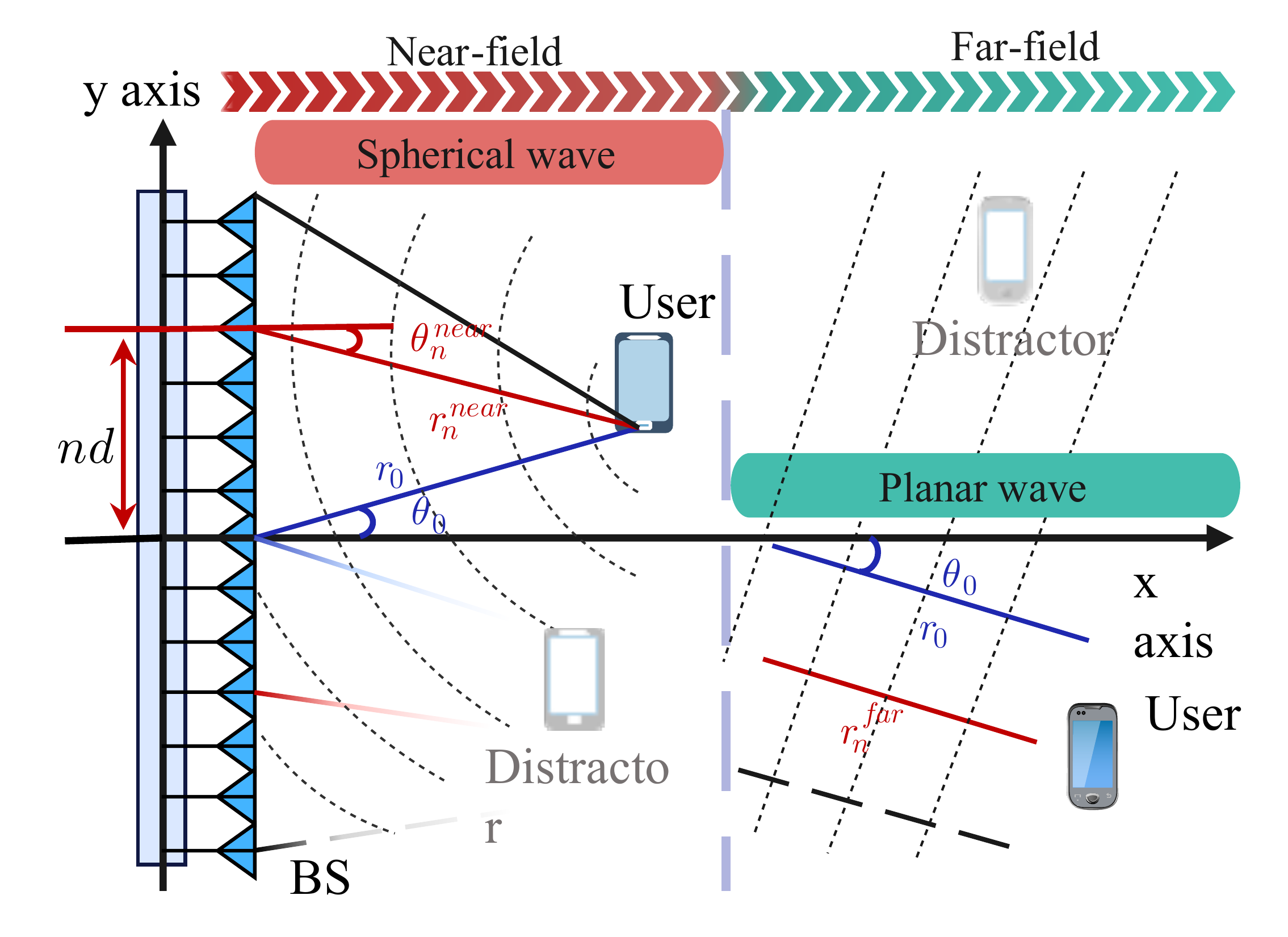}
\caption{Near-field and far-field channel model between BS and UE}
\label{fig2}
\end{figure}

As shown in Fig.~\ref{fig2}, the UE is located at $(r \cos\alpha,r \sin\alpha)$. Considering the characteristic of the near-field spherical wave, the distance between the UE and antenna $n$ is:
\begin{equation}
\label{eq5}
r_{n}=\sqrt{r^{2}+n^{2} d^{2}-2 r n d \sin{\alpha} }.
\end{equation}

Then, the channel model between the $n$-th antenna unit and the UE is expressed by:
\begin{equation}
\label{eq6}
h_{n}=\beta_n e^{-j \frac{2 \pi f_c}{C}(r_n-r)}=\beta_{n}e^{-jM(r_n-r)},
\end{equation}
where $\beta_n=(\frac{\lambda}{4\pi r_n})^2$ is the free space path loss (FSPL) and $M=\frac{2 \pi f_c}{C}$ denotes the wave number. Generally, the distance $r_n$ is much larger than the array aperture $D=(n-1)d$. With $r_n \gg D$, we have $\beta_{-\frac{N}{2}}\approx ,\ldots,\approx \beta_{\frac{N}{2}}\approx \beta =(\frac{\lambda}{4\pi r})^2$. By stacking the channels of all antenna units into one vector, the overall near-field channel vector can be obtained:
\begin{equation}
\label{eq61}
\begin{aligned}
\boldsymbol{h}&=\beta\left[e^{-jM(r_{-\frac{N}{2}}-r)} ,\ldots, e^{-jM(r_{\frac{N}{2}}-r)}\right]\\
&=\sqrt{N}g\boldsymbol{b}(r,\alpha).
\end{aligned}
\end{equation}

It can be observed that the near-field antenna response is a function of the angle $\alpha$ and the distance $r$, and there is no univariate linear correlation. Based on this, we can represent the near-field steering vector as:
\begin{equation}
\label{eq7}
\begin{aligned}
{\boldsymbol{b}}(r,\alpha)=&\left[ e^{-j \frac{2 \pi f_c}{C} ({r_{-\frac{N}{2}}-r}) },\ldots,\right. \\
&\left.e^{j \frac{2 \pi f_c}{C} ({r_n-r}) },\ldots,e^{j \frac{2 \pi f_c}{C} ({r_{\frac{N}{2}}-r})} \right]^{T}.
\end{aligned}
\end{equation}

The received signal of the $k$-th UE can be expressed as:
\begin{equation}
\label{eq8}
y_k=\sqrt{G} \boldsymbol{w}^{H} \boldsymbol{h_k} s+n,
\end{equation}
where $G$ represents transmitting antenna power, $\boldsymbol{w}$ represents the beamforming vector, $\boldsymbol{h_k} $ represents the channel between BS and $k$-th UE, and $s$ is the transmit signal. $n \sim \mathcal{C N}\left(0, \sigma^{2}_n \right)$ represents the additive white Gaussian noise (AWGN). The objective of our beamforming scheme is to maximize the achievable rate of the target UE and minimize interference. For instance, when transmitting signals to UE $1$, the signals received by UE $2$ and UE $3$ are considered interference. Assuming that the sent signal is $|s|^2 =1$, the achievable rate of the $m$-th UE can be expressed as:
\begin{equation}
\label{eq9}
R_k=\log _2\left( 1+\frac{G \left| \boldsymbol{w}^H\boldsymbol{h}_k \right|^2}{\sum_{i\ne k}{\left| \boldsymbol{w}^H\boldsymbol{h}_i \right|}^2+\sigma _{n}^{2}} \right).
\end{equation}

In the above formula, we consider UE interference as noise.

\subsection{Near-field Beamforming}
\begin{figure}[!t]
 \begin{minipage}{0.48\linewidth}
     \vspace{3pt}
     \centerline{\includegraphics[width=0.96\textwidth]{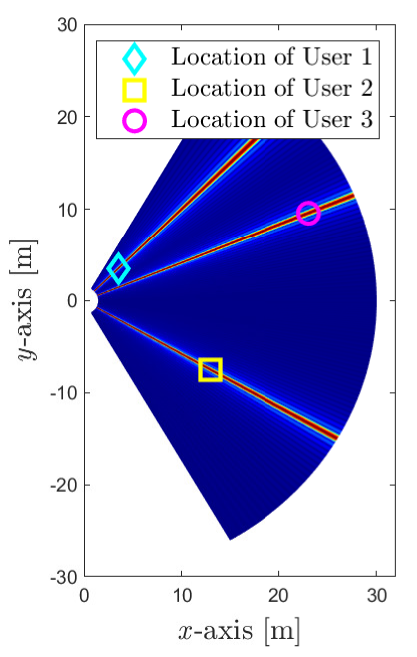}}
     \centerline{(a) Far-field beamforming}
   \end{minipage}
    \begin{minipage}{0.48\linewidth}
     \vspace{3pt}
     \centerline{\includegraphics[width=1.15\textwidth]{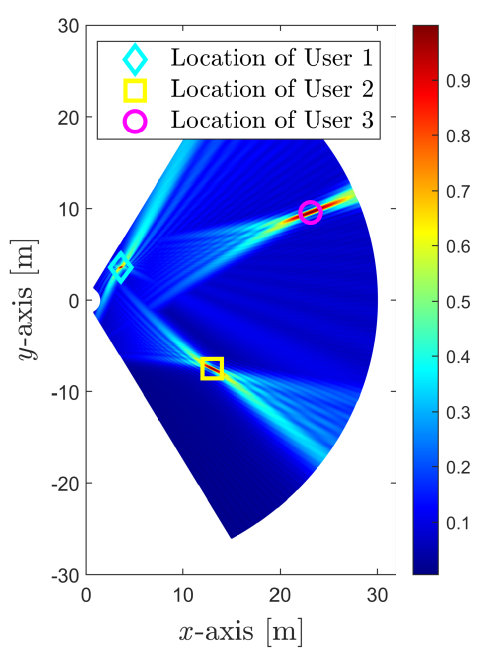}}
     \centerline{(b) Near-field beamforming}
 \end{minipage}
	\caption{(a) represents the standardized signal energy obtained by the receiver through far-field beam focusing, and (b) represents the standardized signal energy obtained by the receiver through near-field beam focusing. The far-field beam travels in different directions, while the near-field beam splitter focuses the beam at different locations.}
	\label{far_near_beam}
\end{figure}

Since the far-field antenna response solely depends on the angle, the far-field beamforming only requires searching in the angle domain. As depicted in Fig.~\ref{far_near_beam} (a), one sector-shaped area can be covered per search. Consequently, this approach introduces significant interference for collinear UEs at different distances but at the same angle. In contrast, the near-field antenna response depends on angle and distance, enabling us to mitigate interference for collinear UEs. As illustrated in Fig.~\ref{far_near_beam} (b), the near-field beam focuses not on a specific angular domain but the UE's location, substantially enhancing the received signal power. However, this advantage comes with the challenge that the near-field codebooks demand more codewords than the far-field codebook. Addressing the challenge of elevated training costs associated with complex codebooks is one of the primary objectives of this study.

We presume the utilization of a predefined codebook for beamforming. The codebook in the polar coordinate domain is expressed as:
\begin{equation}
\begin{aligned}
\mathcal{W}= &\left\{\mathbf{b}\left(\alpha_{1}, r_{1}\right), \ldots,\mathbf{b}\left(\alpha_{1}, r_{j}\right), \ldots, \mathbf{b}\left(\alpha_{1}, r_{J}\right),\right.\\
&\ \ \ \ \ \ \ \ \ \ \ \ \ \ \ \ \ \ \ \ \ \ \left. \ldots \right.\\
& \left.\mathbf{b}\left(\alpha_{i}, r_{1}\right), \ldots,\mathbf{b}\left(\alpha_{i}, r_{j}\right), \ldots, \mathbf{b}\left(\alpha_{i}, r_{J}\right),\right.\\
&\ \ \ \ \ \ \ \ \ \ \ \ \ \ \ \ \ \ \ \ \ \ \left. \ldots \right.\\
&  \left.\mathbf{b}\left(\alpha_{I}, r_{1}\right), \ldots,\mathbf{b}\left(\alpha_{I}, r_{j}\right), \ldots, \mathbf{b}\left(\alpha_{I}, r_{J}\right) \right\},
\end{aligned}
\end{equation}
where each element represents a candidate beam code word, $\alpha_i$ denotes the $i$-th sampling angle, and $r_j$ denotes the $j$-th sampling distance. Generally, the optimal codeword is selected from a predefined codebook to maximize the achievable data rate. Problems related to code word selection can be expressed as:
\begin{equation}
\label{eq11}
\boldsymbol{w}^*=\text{arg}\max_{\boldsymbol{w}\in \mathcal{W}}\log _2\left( 1+\frac{G\left| \boldsymbol{w}^H\boldsymbol{h}_k \right|^2}{\sum_{i\ne k}{\left| \boldsymbol{w}^H\boldsymbol{h}_i \right|}^2+\sigma _{n}^{2}} \right).
\end{equation}

The most straightforward approach is to find the optimal beam through exhaustive beam scanning. However, this method may incur unacceptable training costs compared to far-field scenarios. Moreover, the nonlinear distance information in the near-field channel further complicates beam training, making traditional methods challenging for accurate modeling.

Deep learning excels at dealing with complex nonlinear problems. In the next section, we will explore how to leverage deep learning to accomplish beam training in near-field channels. Our aim is to more effectively capture the channel's nonlinear characteristics and reduce the computational and communication overhead associated with beam training.

\section{Deep learning based near-field beam training Method}
\subsection{Problem Formulation}
\subsection{Problem Formulation}
Near-field beamforming usually has a large codebook size, which means using a codebook-based beam training scheme will incur huge training overhead. In practical scenarios, our primary focus is on the beam gain directed toward the target UE's location. We aim to achieve effective beamforming without relying on a predefined codebook to maximize the achievable data transfer rate. Considering the constant modulus constraint $\lVert w_i \rVert ^2=1, \ \text{for}\,\,i=1,\dotsi, N$, the optimization problem $\boldsymbol{w}$ is given by the following formula:

\begin{equation}
\label{eq12}
\begin{aligned}
\max _ {\left\{{\boldsymbol{w}}\right\}} \ \  &  \sum_{k=1}^{K}R_{k}\left({\boldsymbol{w}}\right) \\
\text { s.t. }  \ \ & \lVert w_i \rVert ^2=1, \forall \ {{i} \in \mathcal{N}= \{1, 2, \ldots, N\}}.
\end{aligned}
\end{equation}

The CSI in XL-MIMO systems is intricate, giving rise to the nonlinearity and significant non-convexity of the aforementioned challenges. 

We introduce deep learning as a powerful tool to address these issues. In this paper, a deep neural network model is designed to flexibly adapt and learn beamforming vectors under complex channel conditions. Our goal is to find the mapping function $\mathbf{\Phi_{\xi}}$ taking $\boldsymbol{\xi}$ as the parameter, which uses CSI to predict the optimal beamforming vector $\boldsymbol{w}$. The mapping function can be expressed as:
\begin{equation}
\mathbf{\Phi_{\xi }} \{ \boldsymbol{h}\} \rightarrow \{  \boldsymbol {\hat {w}}\}, 
\label{eq13}
\end{equation}
where $\boldsymbol{h}$ denotes the downlink CSI from the BS to the UE, and $\boldsymbol{\hat{w}}$ represents the desired beamforming vector.

\subsection{Deep Learning Scheme Design}
\begin{figure}[!t]
\centering
\includegraphics[width=3.6in]{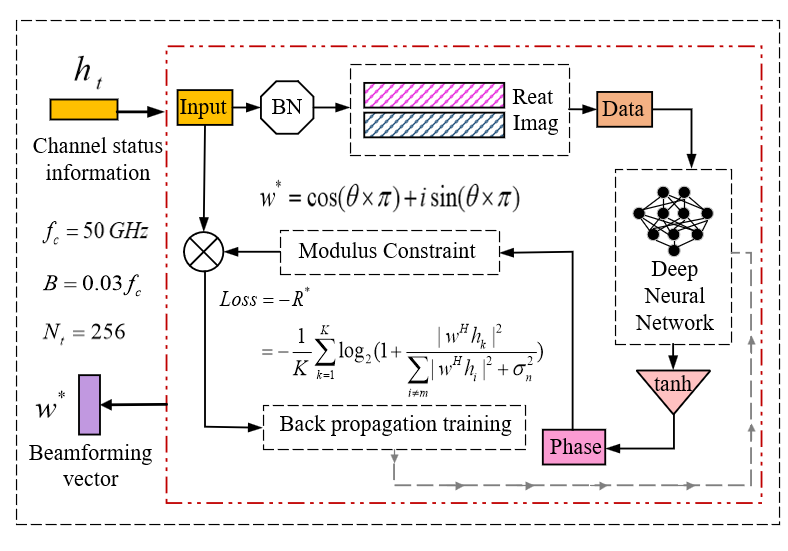}
\caption{Overall architecture diagram of the proposed model.  This shows how to obtain the beamforming vector by using neural network when the channel state information is available.}
\label{dl}
\end{figure}

\begin{algorithm}
\caption{Near-field Beam Training Based on Deep Learning}\label{algorithm}
\KwIn{Channel state information $\boldsymbol{h}$.}
\KwOut{Beamforming vector $\boldsymbol{\hat{w}}$.}
  \textbf{Initialize:} The number of UEs $K$; SNR; The number of transmit antennas $N$; The normalized transmit power $P$\; 
Calculate the input $\boldsymbol{h}$ of neural network according to formula \ref{eq61}.
\textbf{Train the Neural Network:}\;
\For{$epoch \gets 1$ to $E$}
{Randomly sample a batch of training data $(\boldsymbol{h}, \text{SNR})$\;
Calculate the output of the neural network $\boldsymbol{\theta }$\;
Calculate the phase $\boldsymbol{\theta } \leftarrow \pi \boldsymbol{\theta }$\;
Calculate the complex beamforming vector according to formula \ref{eq15}\;
Calculate the loss according to formula \ref{eq16}\;
Update the neural network parameters using backpropagation: $\boldsymbol{\xi} \gets \boldsymbol{\xi} - \eta \cdot \nabla_{\xi} loss$\; 
}
\textbf{Inference:}\;
Given new channel state information $\boldsymbol{h_{\text{new}} }$\; 
Calculate the beamforming vector $\boldsymbol{\hat{w}}=\cos \left( \pi \boldsymbol{\theta } \right) +\text{j}\cdot \sin \left( \pi \boldsymbol{\theta }\right)$\;
\end{algorithm}

The overall architecture of our proposed model is shown in Fig.~\ref{dl}. The channel sample $\boldsymbol{h}$ is input into the neural network. Subsequently, the neural network iteratively learns the model parameters through gradient descent to minimize the predefined loss function. We can converge the model to an approximate global optimal solution by adjusting the learning rate. 

Once the model is trained, all parameters of the network $\xi$ will be fixed. In online deployment, we only need to input the channel sample $\boldsymbol{h}$, and the network will directly output the beamforming vector $\boldsymbol{w}$. Algorithm 1 shows the pseudo-code of our proposed model. In the design of the network structure, we introduce some innovative mechanisms to ensure the effectiveness of the designed model.

\textbf{\textit{1) Input Data Generation:}} Each simulation generates positions for $K$ UEs at $T$ time frames, all of which fall within the specified angle and distance sampling range. When the BS provides service to UE $k$, the remaining $K-1$ UEs are considered as interference. The channel sample $h$ presented in Section IV is used as input data. Detailed parameter design is shown in Table~\ref{t2}.
\begin{table}[H]
\caption{Communication system simulation parameters.}
\centering
\begin{tabular}{c|c|c}
\toprule
Parameter                    & Symbol     & Value \\
\midrule
Number of Transmit Antennas  & $N$      &$256$
\\
Signal Noise Ratio           & SNR        &[$-20$ dB, $20$ dB]\\
Sampling Distance            & $R$        &[$5$ m, $50$ m]\\
Sampling Angle               & $\alpha$   &[$-60$\textdegree, $60$\textdegree]\\
Center Carrier Frequency     & $f_c$      & $50$ GHz\\
Number of UEs                & $K$        & 3\\
Time Frame                   & $T$        & $1000$\\
\bottomrule
\end{tabular}
\label{t2}
\end{table}

\begin{figure*}[!t]
\centering
\includegraphics[width=5.5in]{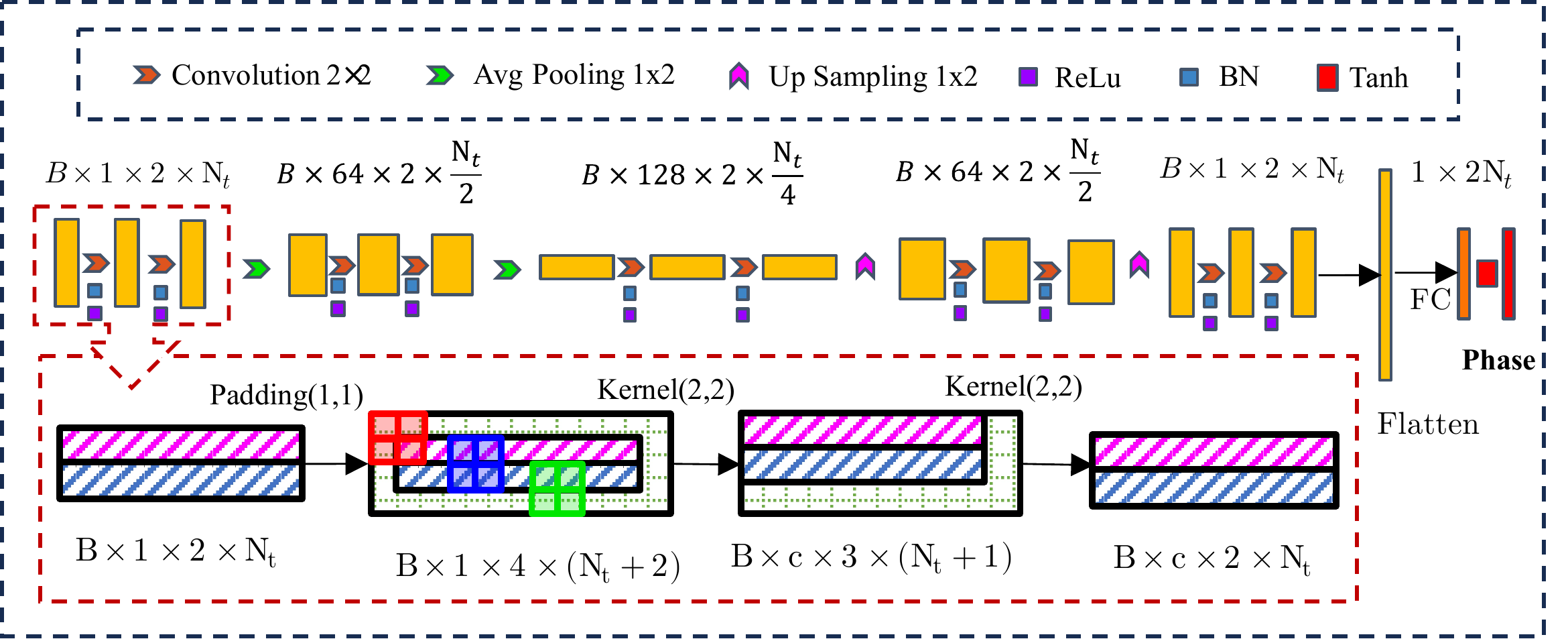}
\caption{Illustrate the structure of the convolutional neural network design. The red dashed box depict the details of the feature extraction block. The orange rectangle represents the data transformations in the convolutional neural network.}
\label{covn}
\end{figure*}

\textbf{\textit{2) Complex Number Feature Extraction: }}
Different from previous approaches that solely rely on fully-connected layers, we leverage convolutional neural networks to extract features from complex signals. The input data dimension is $2\times N$, and we transform it into $1\times2\times N$ by adding an extra dimension. Fig.~\ref{covn} shows the structure of designed convolutional neural networks. A feature extraction block contains two convolution layers, two batch normalization layers (BN), and two rectified linear unit (ReLU). 

By strategically designing the padding and convolution kernel size, we ensure that the data size remains constant before and after passing through a feature extraction block, as shown in the red dashed box in Fig.~\ref{covn}. Initially, convert the input data of size $B\times1\times2\times N$ to $B\times1\times4\times(N+2)$ by adding padding. Then, $2\times2$ convolution layers are used for feature extraction. In the convolution process, the red convolution kernel extracts real part information, the green convolution kernel extracts imaginary part information, and the blue convolution kernel extracts the information combining the real and imaginary parts (phase). After one convolutional layer, the data size becomes $B\times 1\times 3\times (N+1)$. The subsequent convolutional layer restores the data size to $B\times1\times2\times N$, aligning with the dimensions of the input data.

After the data undergoes a feature extraction block, we employ average pooling for downsampling. At this point, the channel count increases, and the data length is halved compared to the original. After repeating this process twice, the data size becomes $B\times C\times 2\times N/4$, where $C$ is the number of channels. Subsequently, after two deconvolution and two feature extraction blocks, the data size becomes $B\times 1\times 2\times N$, matching the dimension of the input data.

\textbf{\textit{3) Convert Angles into Beamforming Vectors: }}
In the left end of Fig.~\ref{covn}, the data is flattened into a one-dimensional vector after the feature extraction block fully captures the features. The data size becomes $1\times N$ following a fully connected layer. To ensure that the neural network's output is a complex-valued vector adhering to the norm constraint, the ``tanh" layer is utilized at the final layer of the network. Specifically, its output is a real value denoted as $\theta$. After passing through the ``tanh" activation layer, its value is confined to the range $(-1, 1)$. Multiplying the output by $\pi$ yields the phase within the range of $[-\pi, \pi]$. The mathematical expression for this process is as follows:
\begin{equation}
\begin{aligned}
\text { Phase }: \theta & \in[-1,1] \\
\theta \times \pi & \in[-\pi, \pi].
\end{aligned}
\label{eq14}
\end{equation}

Then, Euler's formula is applied to convert the phase into the desired complex-valued beamforming vector:
\begin{equation}
\boldsymbol{w}=\exp \left( \text{j}\cdot \pi\boldsymbol{\theta}\right) =\cos \left( \pi \boldsymbol{\theta } \right) +\text{j}\cdot \sin \left( \pi \boldsymbol{\theta } \right) ,
\label{eq15}
\end{equation}
where $j=\sqrt{-1}$. It can be observed that $\theta \pi$ has a clear physical meaning, with its elements corresponding to the equivalent phases of $w$.

\textbf{\textit{4) Loss Function Design: }}
We adopt a codebook-free learning approach, where the training of the neural network is guided by a loss function directly linked to the beamforming target. The loss function for the task is defined as the negative of the average achievable rate.

\begin{equation}
\begin{aligned}
\text { Loss } & = -\frac{1}{Q}\sum_{q=1}^{Q}R^{*} \\
& =-\frac{1}{Q}\sum_{q=1}^{Q} \log _{2}\left(1+\frac{\left|w^{H} h_{k}\right|^{2}}{\sum_{i \neq k}\left|w^{H} h_{i}\right|^{2}+\sigma_{n}^{2}}\right)
\end{aligned}
\label{eq16}
\end{equation}
Where $Q$ represents the total number of training samples and $k$ represents the number of UEs. $\sigma_{n}^{2}$, $h_k$, and $w$ represent the SNR, CSI, and the output analog beamforming vector associated with the $K$th sample, respectively. Notably, the decrease in loss directly corresponds to the increase in the average achievable rate.

\section{Experimental Setup and Simulation Analysis}
\subsection{Experimental Setup}
\begin{table}[H]
\caption{Simulation experiment hyperparameter configuration}
\centering
\begin{tabular}{c|c}
\toprule
Hyperparameter            & Value \\
\midrule
Initial Learning Rate  &$10^{-2}$\\
Batch Size   &$256$\\
Number of Epoch           &$1000$\\
Convolution Kernel Size          &$(2,2)$\\
Optimizer              &Adam\\
\bottomrule
\end{tabular}
\label{t3}
\end{table}

The hyperparameter settings in the proposed deep learning model are shown in Table~\ref{t3}. The data set is randomly divided into a $75\% $ training set and a $25\%$ test set. The batch size is set to $256$. The learning rate is initialized to $10^{-2}$ and dynamically decreases when the evaluation metric is no longer improved. Parameter tuning is conducted using the Adam optimizer.

We utilize Python as the programming language to construct the deep learning network model using the PyTorch framework. The experimental system environment comprises Ubuntu 16.04, Python 3.9, and PyTorch 1.1. The graphics card employed is an NVIDIA GeForce RTX 3060 12GB GPU.



\subsection{Experimental Result}
An evenly spaced half-wave linear array with $Nt = 256$ was deployed at BS during the simulation experiment. The simulation parameters align with the millimeter-wave near-field channel model outlined in Section IV. The average achievable rate is defined as the evaluation indicator. The compared benchmarks are as follows:

\begin{itemize}
\item Near-field hierarchical beam training: The hierarchical near-field codebook comprises several levels of sub-codebooks, which are determined by different sampling angles and distances. We sequentially search from the first-level sub-codebook to the last, where the sampling rate of each subsequent level is influenced by the optimal codeword found in the previous level \cite{WeiCC2022}. Throughout the beam training process, the sampling angles and distances gradually narrow down, ultimately yielding the global optimal codeword.

\item Far-field hierarchical beam training: This is a classical beam training method for far-field scenarios. At the training stage, different levels of codebooks are used for beam alignment, progressively narrowing the beam scanning angle to find the optimal beam direction \cite{NohTWC2017}.

\item Exhaustive search beam training: This scheme simultaneously samples multiple angles $\alpha \in [\alpha_{min},\alpha_{max}]$ and distances $r\in [\rho_{min}, +\infty]$ to construct the near-field codebook. It first fixes the distance loop and exhaustively searches all angles. Then, change the distance loop and repeat the above steps until the entire codebook is traversed to obtain the optimal beam focusing vector \cite{CuiTWC2023}. Exhaustive search beam training incurs significant training overhead, which can be mitigated by reducing the sampling rate but at the cost of reduced beamforming performance. We simulate the exhaustive search beam training schemes with a training cost limit of $256$ and unlimited, respectively.
\end{itemize}

\begin{figure}[!t]
\centering
\includegraphics[width=3.5in]{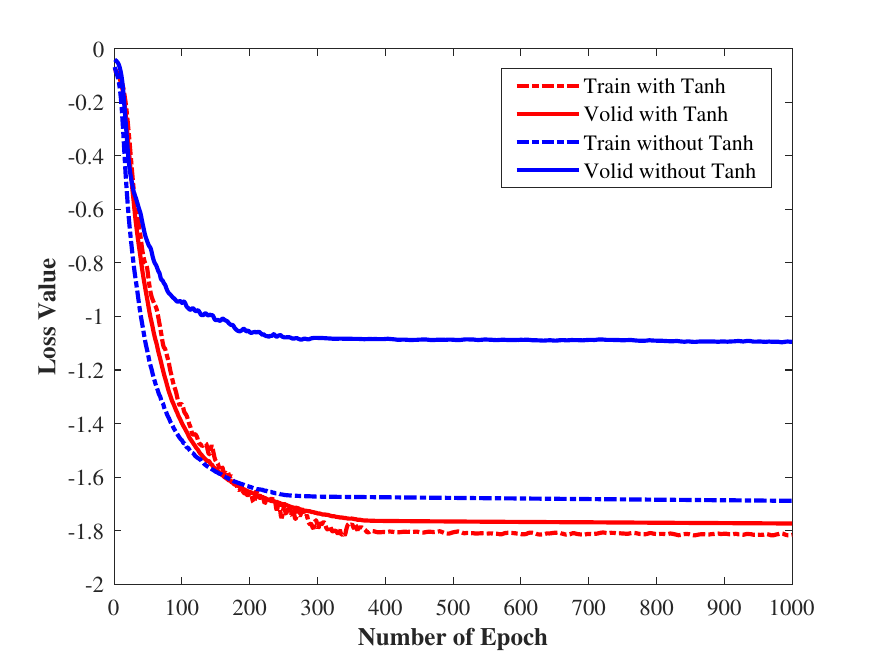}
\caption{Training loss curves of models with and without Tanh layer}
\label{loss}
\end{figure}
The model training loss curve is presented in Fig.~\ref{loss} to validate its effective convergence. These four curves describe the loss evolution of the training and validation sets with and without tanh, respectively. It can be observed that models with tanh layers exhibit superior performance, with lower loss values, and also better performance on verification sets. Conversely, models without tanh layers exhibit pronounced overfitting. This confirms that the incorporation of the tanh layer to confine the phase within the range [-1, 1] enhances the nonlinearity of the model and effectively mitigates overfitting.

\begin{figure}[!t]
\centering
\includegraphics[width=3.5in]{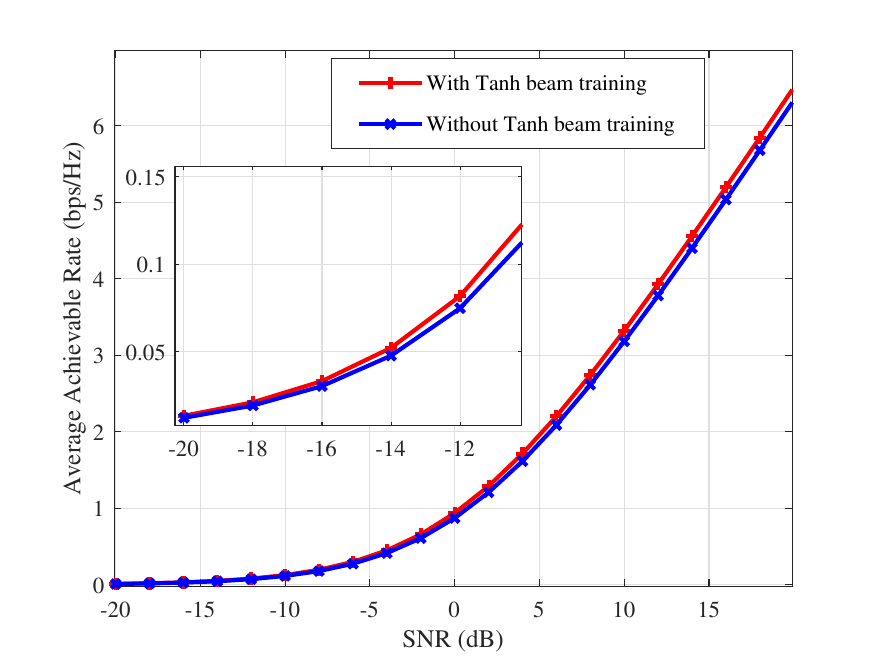}
\caption{Comparison of beamforming effects with and without Tanh layer}
\label{snr_acc}
\end{figure}

We examine whether the inclusion of the tanh layer can enhance the beamforming performance. Fig.~\ref{snr_acc} shows the average achievable rate associated with the beamforming vector output by the model before and after the addition of the tanh layer. The results show that for all SNR values, the model with tanh layer achieves a higher average achievable rate than the model without tanh layer. And this improvement becomes more pronounced with increasing SNR.

\begin{figure}[!t]
\centering
\includegraphics[width=3.5in]{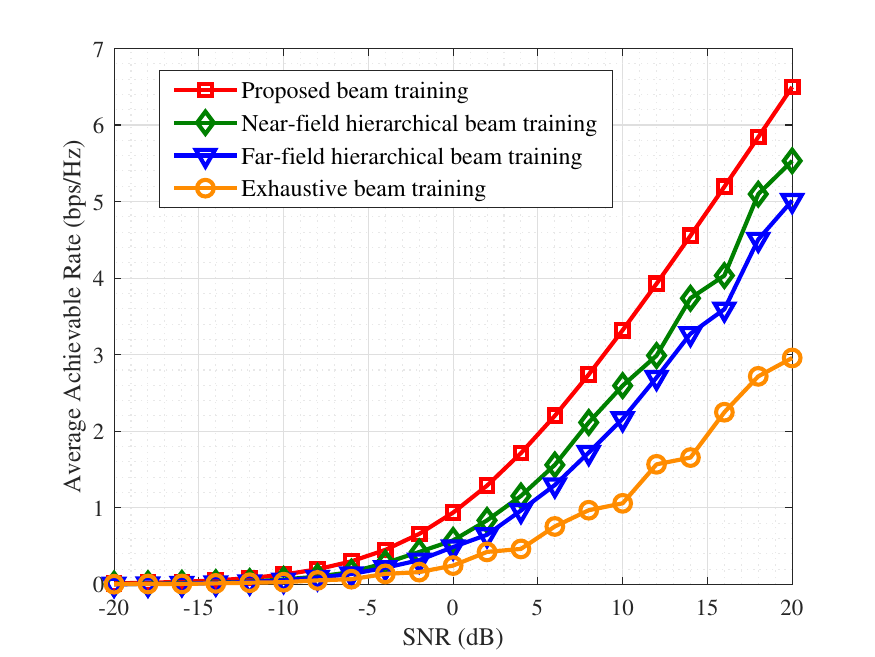}
\caption{The variation curve of the average achievable rate of different schemes with SNR.}
\label{snr_acc_compare}
\end{figure}

Fig.~\ref{snr_acc_compare} shows the variation in the average achievable rate with the SNR under different beam training schemes. It demonstrates that the proposed scheme performs close to the unlimited exhaustive search scheme and consistently outperforms existing far-field and near-field hierarchical beam training schemes across a wide range of SNR values. When SNR = $20$dB, our proposed scheme achieves an approximately $18\%$, $30\%$, and $120\%$ improvement in average achievable rate compared to the near-field hierarchical beam training scheme, far-field hierarchical scheme, and $256$ overhead exhaustive search scheme, respectively. Notably, when the SNR$>-5$dB, the scheme shows an obvious performance gain compared with the comparison scheme. The observed performance enhancement is mainly due to the model aiming to maximize the average achievable rate. The scheme design of codebook-free allows for more dynamic and efficient beamforming compared to traditional methods. Furthermore, the proposed scheme does not exhibit clear superiority at low SNR. Under low SNR conditions, where noise power surpasses signal power, neural networks struggle to effectively extract signal features. Our scheme might be vulnerable to noise under conditions of low SNR. Investigating potential solutions to mitigate noise sensitivity represents a focal point for our future research efforts.

\begin{figure}[!t]
\centering
\includegraphics[width=3.5in]{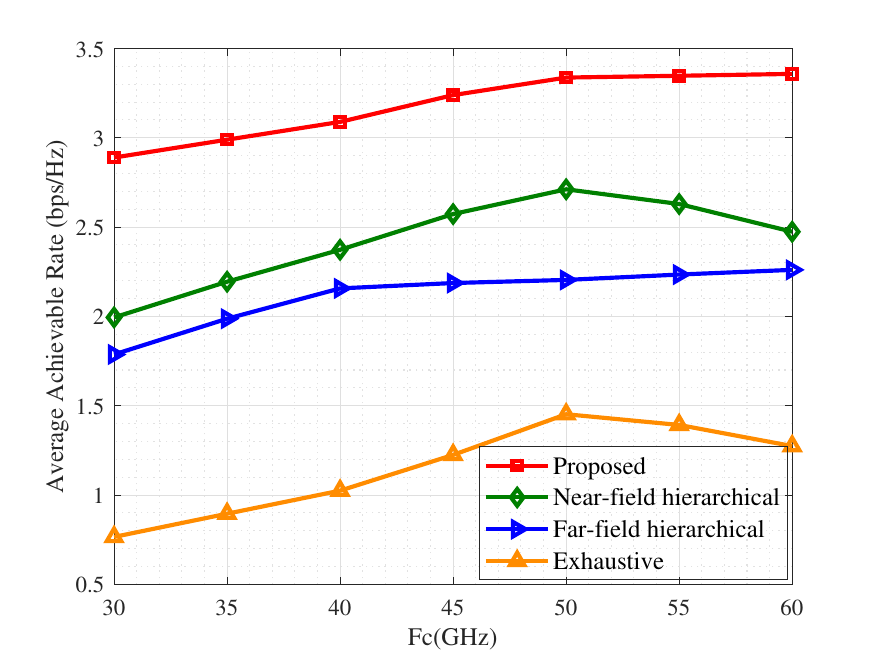}
\caption{The average achievable rate varies with the central carrier frequency.}
\label{fc}
\end{figure}

Carrier frequency is a crucial factor influencing communication rate. Fig.~\ref{fc} shows the average achievable rate curves of the proposed scheme and the compared scheme at different carrier frequencies. The figure illustrates that for all carrier frequencies, the average achievable rate of our proposed scheme consistently outperforms other schemes. With the increase of carrier frequency, near-field propagation gradually dominates, resulting in the average achievable rate of the far-field scheme no longer increasing when the carrier is greater than $40$ GHz. Optimal performance for all solutions is observed at the carrier frequency of $50$ GHz. This is attributed to the increased FSPL of high-frequency signals during transmission. As the carrier frequency increases, the gain from increasing the number of antennas will not be enough to offset FSPL. Notably, when $f_c >$ $50$ GHz, the performance of the proposed scheme performance remains stable, affirming the robustness of our proposed solution. Given the overhead of the exhaustive search scheme is limited to $256$, the scheme encounters challenges in conducting precise searches in high-frequency bands, leading to a notable degradation in performance.

\begin{figure}[!t]
 \begin{minipage}{0.9\linewidth}
     \vspace{3pt}
     \centerline{\includegraphics[width=1.1\textwidth]{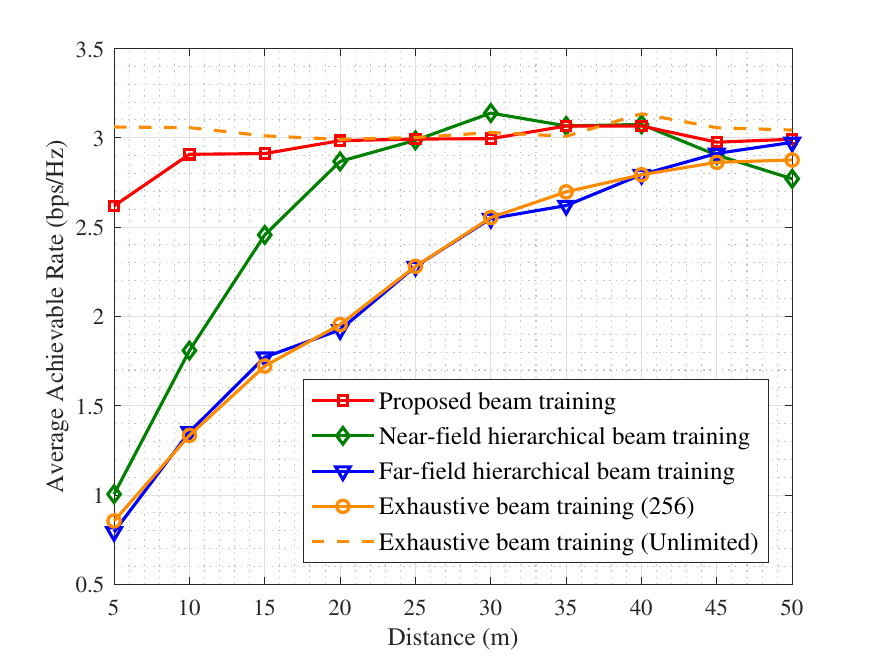}}
     \centerline{(a)}
     \vspace{3pt}
     \centerline{\includegraphics[width=1.1\textwidth]{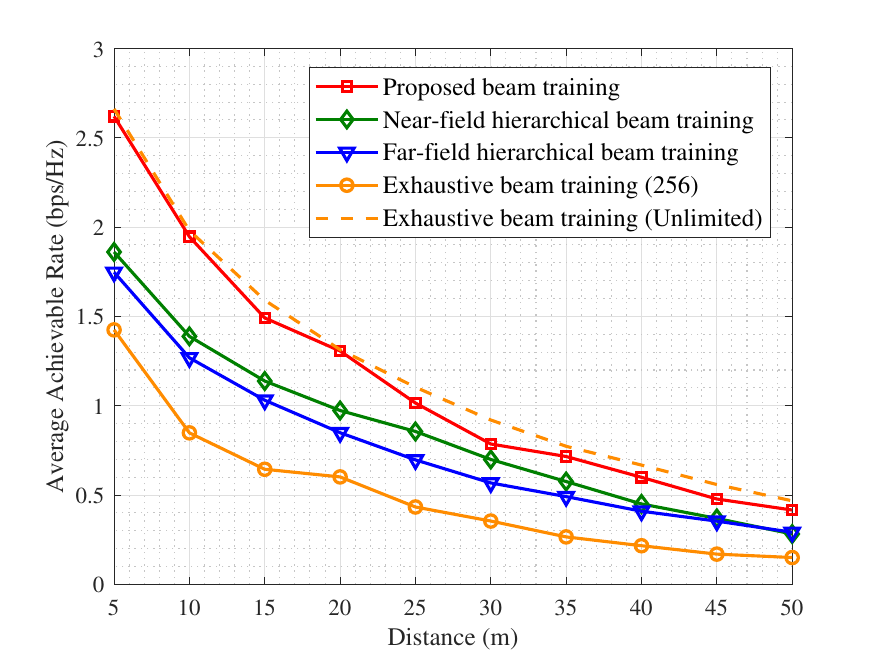}}
     \centerline{(b)}
 \end{minipage}
	\caption{(a) and (b) represent the variation of the average reachable rate with user distance without and with path loss, respectively.}
	\label{distance}
\end{figure}

Fig.~\ref{distance} illustrates the average achievable rate against the UE-to-BS distance. Here, the distance $r$ from the UE to the BS gradually increases from $5$ to $50$ meters, and the SNR is set to $10$ dB. As can be seen from Fig.~\ref{distance} (a), when FSPL is not considered, the average achievable rate of our proposed scheme is superior to all the compared schemes in the near-field region. As the distance increases, the performance of our proposed scheme is comparable to that of the far-field hierarchical beam training scheme. Since only angle search can be performed, the performance of far-field hierarchical schemes deteriorates rapidly in the near-field region. Due to the maximum training cost limitation, the $256$ exhaustive scheme also experiences severe degradation in the near-field region, impeding the practical search for near-field positions. Additionally, the near-field hierarchical beam training scheme can moderately alleviate the decline in the average achievable rate, but its performance gradually diminishes when the distance exceeds $30$ meters. Fig.~\ref{distance} (b) shows that due to FSPL, the performance of all beam training schemes decreases as the distance increases. However, the proposed beam training scheme still maintains high performance at all distances and is close to the performance of the exhaustive scheme without limited training cost. In contrast, our proposed scheme demonstrates the ability to search for the optimal beamforming vector with minimal pilot overhead. This affirms the robustness of the proposed scheme in both near and far-field communications scenarios.

\begin{figure}[!t]
\centering
\includegraphics[width=3.5in]{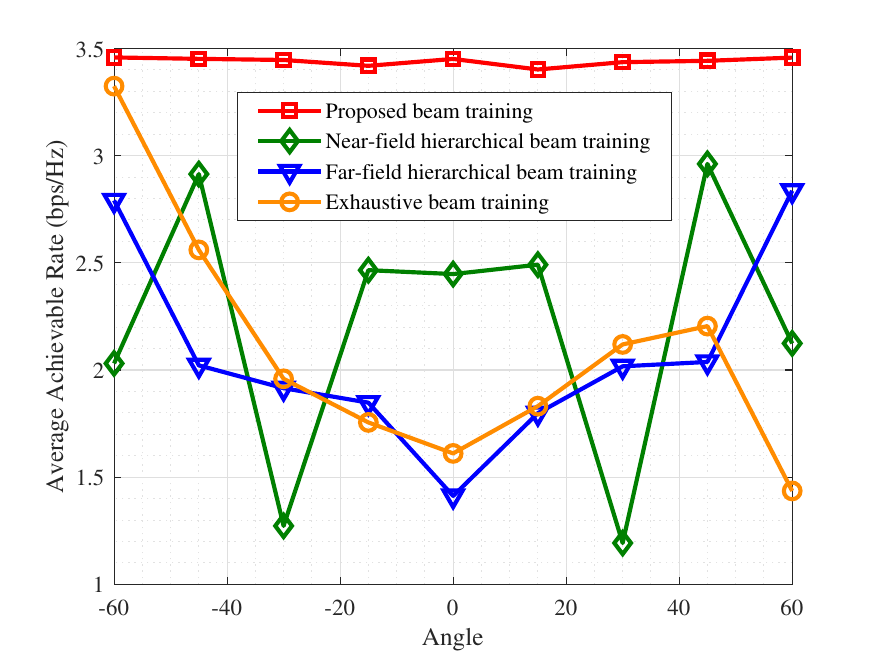}
\caption{Average achievable rate curves at different angles.}
\label{angle}
\end{figure}

Fig.\ref{angle} depicts the variation in average achievable rate with angle $\alpha$. Here, $\alpha$ gradually increases from $-60$\textdegree to $60$\textdegree, and the SNR is set to $10$ dB. Notably, for the far-field scheme, performance degrades significantly as $\alpha$ approaches $0$. This degradation is attributed to the increased prominence of near-field effects near zero angles, as demonstrated in the literature \cite{Cui2021}. Furthermore, the near-field hierarchical beam training scheme exhibits noticeable fluctuations. This is because the near-field hierarchical scheme creates the codebook by uniformly sampling angles and distances in cartesian coordinates. This approach proves challenging in achieving stable beamforming performance across the entire near-field environment \cite{CuiTCOM2022}. In contrast, our proposed scheme consistently attains the high average achievable rate across all angles. This underscores the robustness and effectiveness of our approach in handling the challenges posed by near-field communications scenarios.

\begin{figure}[!t]
\centering
\includegraphics[width=3.5in]{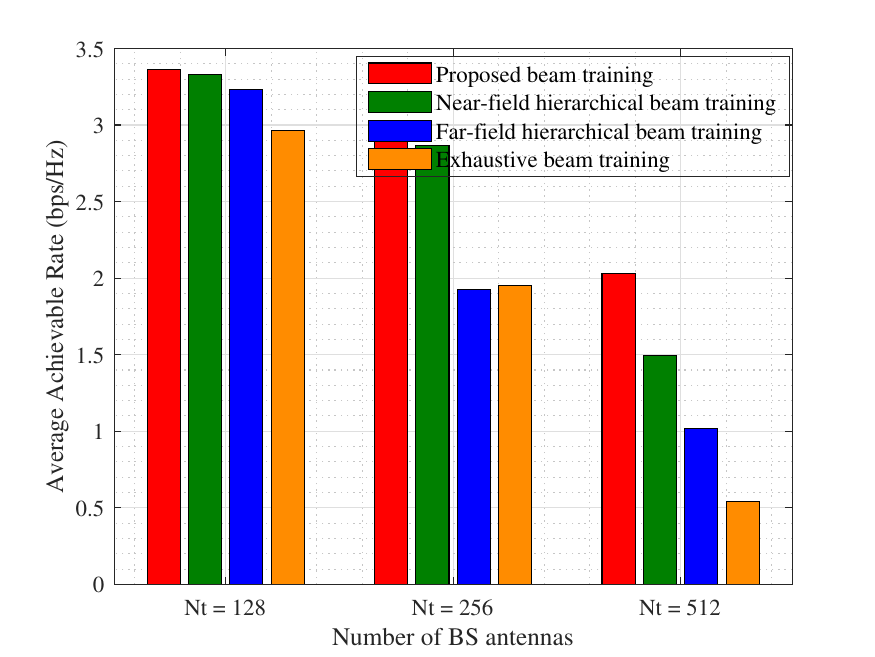}
\caption{Impact of transmitting antenna quantity on beamforming performance.}
\label{nt}
\end{figure}


Fig.~\ref{nt} illustrates the average achievable rate under different numbers of BS antennas $N$. We can see that the performance of each scheme tends to decline with the increase in the number of antennas. Remarkably, our proposed scheme consistently exhibits superior compared to all comparison schemes. With the growth in the number of antennas, the performance of the proposed model improves more significantly than the compared model. The escalating number of antennas makes it more difficult for traditional schemes to deal with intricate beam codebooks. The consistent outperformance of our model in the face of increasing antenna numbers underscores its adaptability and suitability for evolving communication scenarios.

\section{Conclusions}

In this paper, we proposed a near-field beamforming scheme based on deep learning. By strategically designing the padding and kernel size of the convolutional neural network, we effectively extract the features of the complex CSI. During the network training process, we utilize the negative value of the total achievable rate in the multi-user network as the loss function and ensure the amplitude constraint of the beamforming vector through a customized Tanh layer. Our solution doesn't rely on predefined codebooks and only requires the estimated CSI as input to obtain the optimal beamforming vector. Simulation results demonstrate the competitive performance of the proposed scheme.


\begin{IEEEbiography}[{\includegraphics[width=1in,height=1.25in,clip,keepaspectratio]{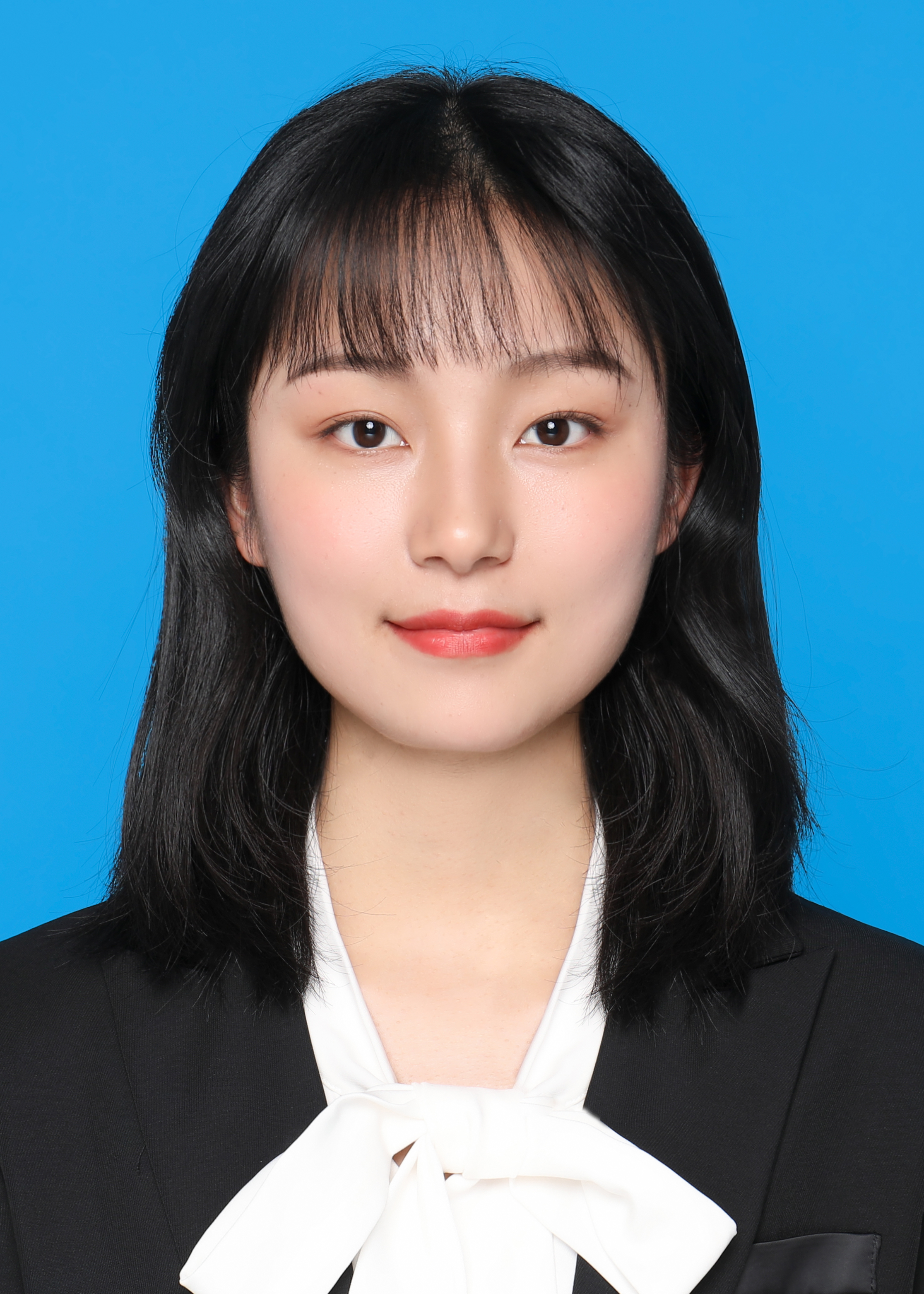}}]{Jiali Nie}
 is currently studying for a master’s degree in the Beijing University of Posts and Telecommunications (BUPT), Beijing 100876, China. Her main research interests include wireless communication, integrated sensing and communication (ISAC), intelligent signal processing, and deep learning. She served as a reviewer for several journals such as IEEE Communication Magazine. She also served as a Technical Program Committee (TPC) Member for several conference.
\end{IEEEbiography}

\begin{IEEEbiography}[{\includegraphics[width=1in,height=1.25in,clip,keepaspectratio]{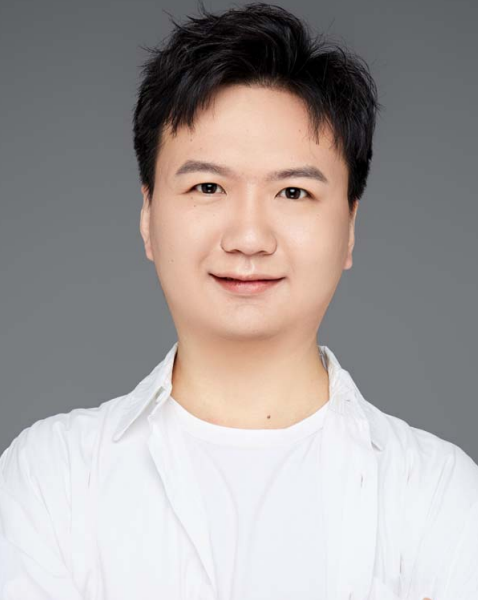}}]{Yuanhao Cui}
(Member, IEEE) received the B.Eng. degree (Hons.) from Henan University and the Ph.D. degree from the Beijing University of Posts and Telecommunications (BUPT). He has been the CTO and the Co-Founder of two startup companies, where he has invested more than $\$3$ million dollars. He holds more than 20 granted patents. His research interests include precoding and protocol designs for ISAC. He is a member of the IMT-2030 (6G) ISAC Task Group. He received the Best Paper Award from IWCMC 2021. He is the Founding Secretary of the IEEE ComSoc ISAC Emerging Technology Initiative (ISAC-ETI) and the CCF Science Communication Working Committee. He was the Organizer and the Co-Chair for a number of workshops and special sessions in flagship IEEE conferences, including ICC, ICASSP, WCNC, and VTC. He is the Lead Guest Editor for the Special Issue on Integrated Sensing and Communication for 6G of the IEEE OPEN JOURNAL OF THE COMMUNICATIONS SOCIETY and the Guest Editor for the Special Issue on Integrated Sensing and Communications for Future Green Networks of the IEEE TRANSACTIONS ON GREEN COMMUNICATIONS AND NETWORKING.
\end{IEEEbiography}

\begin{IEEEbiography}[{\includegraphics[width=1in,height=1.25in,clip,keepaspectratio]{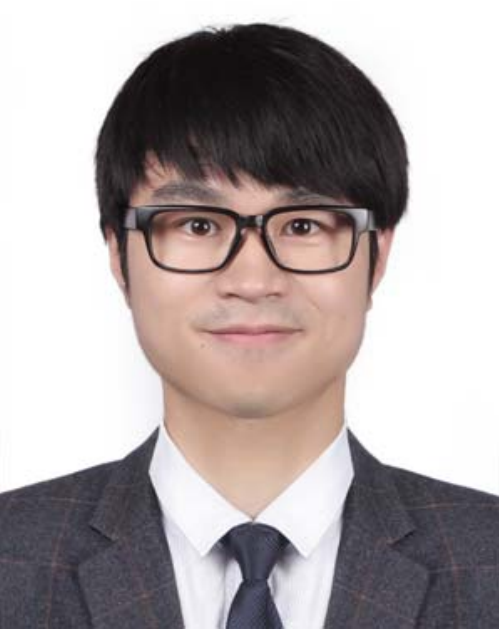}}]{Zhaohui Yang} (Member, IEEE) received the Ph.D. degree from Southeast University, Nanjing, China, in 2018. From 2018 to 2020, he was a Postdoctoral Research Associate with the Center for Telecommunications Research, Department of Informatics, King’s College London, London, U.K. From 2020 to 2022, he was a Research Fellow with the Department of Electronic and Electrical Engineering, University College London, London. He is currently a ZJU Young Professor with the Zhejiang Key Laboratory of Information Processing Communication and Networking, College of Information Science and Electronic Engineering, Zhejiang University, Hangzhou, China, and also a Research Scientist with the Zhejiang Laboratory. His research interests include joint communication, sensing, computation, federated learning, and semantic communication. He was the recipient of the 2023 IEEE Marconi Prize Paper Award, 2023 IEEE Katherine Johnson Young Author Paper Award, 2023 IEEE ICCCN Best Paper Award, and the first Prize in Invention and Entrepreneurship Award of the China Association of Inventions. He was the Co-Chair of international workshops with more than ten times, including IEEE ICC, IEEE GLOBECOM, IEEE WCNC, IEEE PIMRC, and IEEE INFOCOM. He is an Associate Editor for the IEEE COMMUNICATIONS LETTERS, IET Communications, and EURASIP Journal on Wireless Communications and Networking. He wast the Guest Editor for several journals, including IEEE JOURNAL ON SELECTED AREAS IN COMMUNICATIONS.
\end{IEEEbiography}

\begin{IEEEbiography}[{\includegraphics[width=1in,height=1.25in,clip,keepaspectratio]{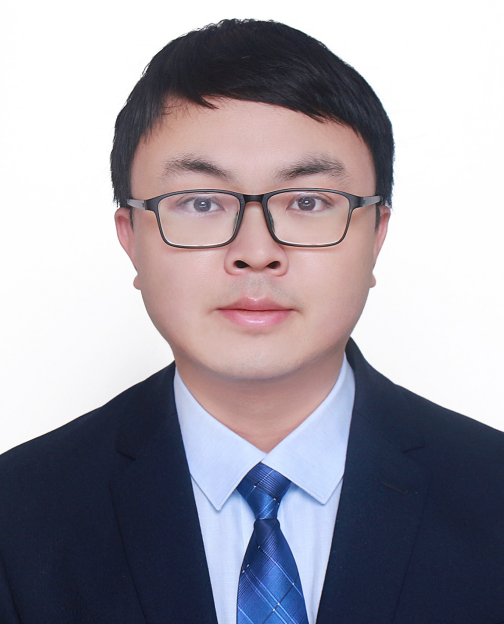}}]{Weijie Yuan}
(Member, IEEE) received the B.E. degree from the Beijing Institute of Technology, Beijing, China, in 2013, and the Ph.D. degree from the University of Technology Sydney, Sydney, NSW, Australia, in 2019. From 2019 to 2021, he was a Research Associate with the University of New South Wales, Sydney, NSW, Australia. He is currently an Assistant Professor with the School of System Design and Intelligent Manufacturing, Shenzhen, China. He was a Research Assistant with the University of Sydney, Sydney, NSW, Australia, a Visiting Associate Fellow with the University of Wollongong, Wollongong, NSW, Australia, and a Visiting Fellow with the University of Southampton, Southampton, U.K., from 2017 to 2019. In 2016, he was a Visiting Ph.D. Student with the Institute of Telecommunications, Vienna University of Technology, Austria. His research interests include statistical signal processing, OTFS, and ISAC. He is the CoChair and Co-Organizer for workshops and special sessions on orthogonal time frequency space (OTFS), and integrated sensing and communication (ISAC) in ICC 2021, ICCC 2021, SPAWC 2021, VTC 2021-Fall, WCNC 2022, and ICASSP 2022. He is the Founding Chair of the IEEE ComSoc special interest group on OTFS (OTFS-SIG) and is serving as an Associate Editor for the EURASIP Journal on Advances in Signal Processing.
\end{IEEEbiography}

\begin{IEEEbiography}[{\includegraphics[width=1in,height=1.25in,clip,keepaspectratio]{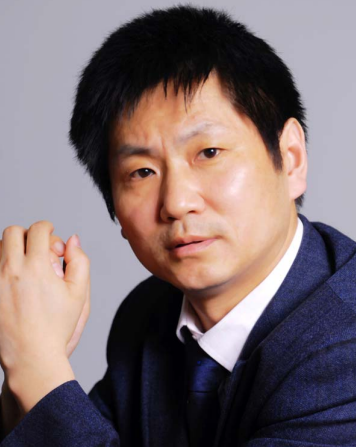}}]{Xiaojun Jing}
(Member, IEEE) received the M.S. and Ph.D. degrees from the National University of Defense Technology in 1995 and 1999, respectively. He is currently a Professor at the Beijing University of Posts and Telecommunications. His research interests include wireless communication, information security, and image processing.
\end{IEEEbiography}

\end{document}